# Towards Intelligent Urban Park Development Monitoring: LLM Agents for Multi-Modal Information Fusion and Analysis

Zixuan Xiao*
*Department of Urban Planning and Design*
*The University of Hong Kong*
Hong Kong SAR
zxxiao@connect.hku.hk

Chunguang Hu
*School of Urban Planning and Design*
*Peking University*
Shenzhen, China
huchunguang@stu.pku.edu.cn

Jun Ma
*Department of Urban Planning and Design*
*The University of Hong Kong*
Hong Kong SAR
junma@hku.hk

*Abstract*—As an important part of urbanization, the development monitoring of newly constructed parks is of great significance for evaluating the effect of urban planning and optimizing resource allocation. However, traditional change detection methods based on remote sensing imagery have obvious limitations in high-level and intelligent analysis, and thus are difficult to meet the requirements of current urban planning and management. In face of the growing demand for complex multi-modal data analysis in urban park development monitoring, these methods often fail to provide flexible analysis capabilities for diverse application scenarios. This study proposes a multi-modal LLM agent framework, which aims to make full use of the semantic understanding and reasoning capabilities of LLM to meet the challenges in urban park development monitoring. In this framework, a general horizontal and vertical data alignment mechanism is designed to ensure the consistency and effective tracking of multi-modal data. At the same time, a specific toolkit is constructed to alleviate the hallucination issues of LLM due to the lack of domain-specific knowledge. Compared to vanilla GPT-4o and other agents, our approach enables robust multi-modal information fusion and analysis, offering reliable and scalable solutions tailored to the diverse and evolving demands of urban park development monitoring.

## I. Introduction

As a key element of modern urbanization, urban parks have a profound impact on the ecological environment of cities and the quality of life of urban residents [1]. With the acceleration of urbanization, the number of newly constructed parks continues to increase. How to efficiently and accurately monitor the development process of these parks has become an important issue in the field of urban planning and management. Traditional monitoring methods mainly rely on remote sensing images and change detection methods. However, these methods often lack semantic interpretation of changes and are not flexible enough to adapt to diverse analysis needs.

Change detection methods, whether based on traditional statistical analysis or deep learning approaches, focus on quantitative analysis of pixel-level changes [2]. For example, these methods can provide increasingly accurate change maps, but these maps cannot meet higher-level analysis needs, unable to provide a comprehensive analysis that combines qualitative and quantitative analysis, such as when and what facilities were built during the development of the park, the change in the proportion of land use types before and after the construction of the park, and whether the green area of the park has increased after the construction. These analyses are crucial to understanding the impact of park development, evaluating planning effects, and optimizing resource allocation.

Recently, large language models (LLMs) have demonstrated remarkable abilities in semantic understanding and reasoning, which makes them a potential tool for intelligent change analysis and development monitoring [3]. Specifically, LLMs can perform advanced analysis of complex problems and adapt to different application scenarios to meet the growing demand for intelligent analysis. However, LLMs also face challenges in application, especially in scenarios where they lack of domain-specific knowledge, and are prone to hallucination issues, generating inaccurate or unreliable results. In addition, in multimodal data analysis, there is a lack of semantic alignment mechanism between input and output, resulting in insufficient data fusion and consistency.

To solve these problems, this paper proposes an LLM agent framework that provide multi-modal information fusion and analysis. To support global multi-modal data alignment, a generalized data alignment mechanism is designed to ensure the integrity of the multi-modal data analysis and solve the consistency problem in multi-modal data processing. Moreover, a specialized toolkit is constructed to alleviate the hallucination issues.

In conclusion, this work introduces a novel framework for monitoring the development of urban parks, offering new perspectives and practical solutions that position urban park monitoring as a crucial element in smart city initiatives. The main contributions of this research are as follows:

1. A multi-modal LLM agent framework is proposed for intelligent urban park development monitoring, leveraging the advanced semantic understanding and reasoning capabilities of LLM agents. This framework enables high-level change analysis, such as qualitative and quantitative evaluations, addressing the limitations of traditional change detection methods in intelligence and analysis depth.

2. A generalized data alignment mechanism is designed to address the complexity of multi-modal data integration while ensuring consistency and traceability throughout the entire data analysis workflow, effectively aligning user semantics with the system.

3. A specialized toolkit is developed to mitigate hallucination issues in LLMs caused by a lack of domain-specific knowledge, enabling reliable and efficient analysis.

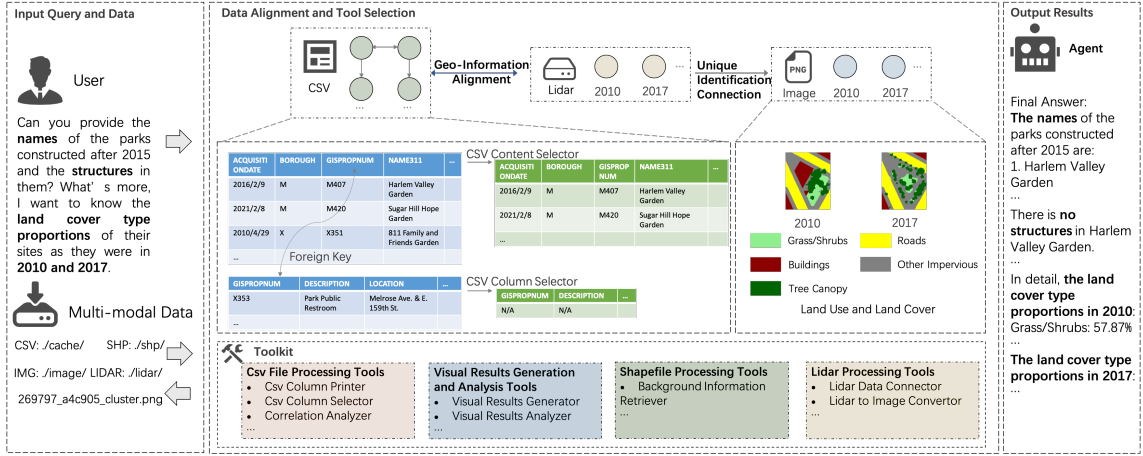

Fig. 1. The overall framework of our multi-modal agent for intelligent urban park development monitoring.

## II. RELATED WORK

Remote sensing change detection methods can be broadly categorized into traditional and deep learning-based approaches. Traditional methods, including algebra-based, statistics-based, and transformation-based techniques, are computationally efficient and widely used. However, they often rely on manual parameter tuning and are limited in handling complex datasets, making them less effective for high-level analysis [4].

Deep learning-based methods, such as Post-classification Change Methods, Differencing Methods, Direct Classification Methods, and Differencing Neural Network Methods, leverage the ability to learn complex features directly from data [5]. These methods significantly improve detection accuracy and adaptability. Advanced models, including recurrent neural networks and adversarial methods, further enhance performance by addressing temporal dependencies and noise.

Despite their strengths, both traditional and deep learning methods primarily rely on single analytical approaches, limiting their ability to perform high-level qualitative and quantitative analyses across diverse data types.

With the rise of Large Language Models (LLMs) and their exceptional few-shot capabilities compared to deep learning models [6, 7], prompting has become a key technique for guiding these models toward specific outputs. Methods like Chain of Thought (CoT) and In-Context Learning (ICL) further enhance LLM performance by enabling logical reasoning and learning from examples directly within the prompt, without retraining [8, 9].

As intelligent systems evolve, multi-modal agents have gained prominence by integrating diverse data types, such as text, images, and audio, into unified frameworks. These agents utilize pre-trained encoders to adapt various modalities for LLM processing. While systems like SEED-LLaMA and SpeechGPT demonstrate the potential of such integration, they require extensive multi-modal training data and significant computational resources [10, 11].

To address these challenges, tool-assisted agent construction has emerged as a practical alternative, connecting LLMs with external tools to perform specialized tasks without extensive retraining [12]. Our framework introduces a generalizable approach for constructing multi-modal agents tailored to urban park development monitoring, reducing hallucination by integrating domain-specific tools and methods.

## III. METHOD

To support intelligent urban park development monitoring, a multi-modal agent framework is proposed. As illustrated in Fig. 1, this framework processes input queries with multi-modal data in two main steps. First, after the agent decompose the input task with its reasoning ability into different subtasks, the agent needs to perform data alignment for multi-modal data input. Second, the agent select the relevant tool for these decomposed subtasks. In the data alignment stage, the agent ensures the consistency of different modal data in space and time for effective analysis. In the tool selection stage, the agent selects appropriate tools for data processing and analysis according to the specific requirements of the subtask. After these two steps, the agent integrates the results of each subtask and outputs it as a complete analysis report to provide users with comprehensive monitoring results and decision support.

The framework aims to provide intelligent monitoring of the development process of urban parks and comprehensive analytical results through the integration and analysis of multi-modal data. The development of urban parks involves many aspects, such as green space construction, infrastructure layout, and ecological environment improvement. In order to comprehensively monitor these aspects, the framework needs to process multiple types of data, including remote sensing images, geographic information system (GIS) data, socio-economic data, etc. These data reflect the development status and environmental changes of urban parks from different perspectives. Therefore, data alignment is a key step to ensure the consistency and effective analysis of multi-modal data.

In this framework, we design a general horizontal and vertical data alignment mechanism. Horizontal data alignment is mainly used for alignment between multi-modal data. In urban park development monitoring, data of different modalities often have different spatial resolutions and temporal frequencies. For example, the spatial resolution of remote sensing images may be high, which can clearly show the details of the park, but the temporal frequency may be low, which can only reflect the state of the park at a specific time point. While the temporal frequency of tabular data is high, which can reflect the descriptive conditions of the park in real time. In order to achieve effective fusion and analysis of these data, the framework ensures accurate spatial correspondence of data through geo-information alignment between different modal data. Specifically, the framework converts geo-information in each modal data into the coordinates in the same coordinate system. This horizontal data alignment mechanism can provide a comprehensive analytical perspective, making the analysis results more accurate and reliable.

Vertical data alignment is used to record the relationship before and after data processing. In the process of urban park development monitoring, data needs to go through multiple processing steps in subtasks. In order to ensure the consistency and traceability of data during the processing process, the framework assigns a global unique identifier to each data element. This identifier will run through the entire process of data processing, so that the state and changes of the data in each processing step can be accurately recorded and tracked. Through vertical data alignment, the relationship between the processed data and the original data can be ensured, which improves the reliability of the system and ensures the semantic alignment between the user and the system. Moreover, this data consistency mechanism provides the interpretability of the whole system through the data tracking.

Another important part of the system is the toolkit. In order to solve the hallucination issues that may occur in the application of LLM, we provide corresponding processing tools for different data types to integrate professional knowledge information in the field. For example, for CSV data, we developed a CSV Column Selector to filter column data in the table. In urban park development monitoring, CSV tabular data is often used to store park attribute information, construction progress data, and statistical data of surrounding areas. Through the CSV column selector, users can select specific columns for analysis as needed, such as selecting columns such as the construction area and investment amount of the park to more accurately analyze the construction status and economic benefits of the park. The design of this toolkit enables our agent to process and analyze various data types more accurately, thereby improving the reliability and accuracy of the analysis results.

The proposed framework employs a decision-level fusion strategy to integrate heterogeneous data sources, including tabular data and geospatial information derived from LiDAR and remote sensing. Each modality is processed independently using dedicated tools. In particular, LiDAR data is converted into remote sensing imagery for computing land use and land cover (LULC) proportions. The final integration is performed by a large language model (LLM), which reasons over the structured outputs to generate task-specific decisions.

This multi-modal framework enables intelligent monitoring of urban park development, providing a data-driven basis for evaluating progress and optimizing resource allocation. By combining diverse data sources through agent-based processing and LLM reasoning, it supports more accurate and sustainable urban planning.

## IV. Experiment

We constructed a question dataset to evaluate the performance of various models and agents in urban parks development analysis. The dataset includes questions spanning three levels of analysis, basic, qualitative, and quantitative. The data source of the urban parks are from NYC open data (https://opendata.cityofnewyork.us). Basic questions focus on straightforward data retrieval and description, while qualitative questions require higher-level interpretation, such as summarizing public structures development patterns in urban parks. Quantitative questions are the most complex, involving numerical evaluations, such as land use and land cover changes. TABLE I provides representative examples for each question type.

TABLE I. Examples from question dataset.

| Analysis Level | Number | Example |
| --- | --- | --- |
| Basic | 4 | Can you tell me the names of the parks constructed in 2015 that are smaller than 2 acres? |
| Qualitative | 2 | Please provide the names, addresses, and boroughs of the parks constructed in 2015, along with the number and locations of drinking fountains inside the parks, if any |
| Quantitative | 4 | Please provide the names of the parks constructed in December 2017 and the land cover type proportions of their sites as they were in 2010. |

We construct our agent with GPT-4o as LLM backend. As result, for benchmarking, we tested our proposed agent against several other models and agents. These included vanilla GPT-4o, the LangChain default agents (SQL, pandas DataFrame, and CSV agents), and a single-modality agent designed specifically for handling CSV data. Vanilla GPT-4o served as a baseline general-purpose LLM without domain-specific adaptation, while the LangChain agents represented agent with default data processing tools (https://www.langchain.com/agents). The single-modality CSV agent was built with multiple CSV tools effectively but lacked multi-modal integration capabilities.

TABLE II. Performance on question dataset of different models. Note: A check mark (✓) indicates a correct answer, while a cross (✗) indicates an incorrect one.

| Analysis Level | Models \ Questions | LangChain Default Agent | Vanilla GPT-4o | Single-modality Agent | Our Agent |
|---|---|---|---|---|---|
| Basic | Q1 | ✗ | ✓ | ✓ | ✓ |
| Basic | Q2 | ✗ | ✓ | ✓ | ✓ |
| Basic | Q3 | ✗ | ✓ | ✓ | ✓ |
| Basic | Q4 | ✗ | ✓ | ✓ | ✓ |
| Qualitative | Q5 | ✗ | ✗ | ✓ | ✓ |
| Qualitative | Q6 | ✗ | ✗ | ✓ | ✓ |
| Quantitative | Q7 | ✗ | ✗ | ✗ | ✓ |
| Quantitative | Q8 | ✗ | ✗ | ✗ | ✓ |
| Quantitative | Q9 | ✗ | ✗ | ✗ | ✓ |
| Quantitative | Q10 | ✗ | ✗ | ✗ | ✓ |

The results, summarized in TABLE II, reveal distinct performance differences across models. The LangChain default agents, including the SQL Agent, Pandas DataFrame Agent, and CSV Agent, offer a straightforward way to interact with structured data using predefined processing tools. However, their rigid workflows and general-purpose tools limit their adaptability and effectiveness for complex data processing tasks. A key limitation is their prompt construction strategy, where data tables are directly embedded into prompts. This often leads to token limit issues, causing errors when dealing with large datasets. Additionally, these agents lack data lineage tracking, making it difficult to handle multi-step transformations while preserving data integrity. This can result in data contamination or loss, particularly in scenarios requiring extensive workflows. In our experiments, these limitations were evident. The LangChain agents struggled with large datasets and complex multi-modal queries.

Vanilla GPT-4o performed adequately for basic questions, successfully retrieving data from individual CSV files. However, it struggled with qualitative and quantitative tasks, failing to handle foreign key identification or multi-modal data integration, often producing hallucinated or irrelevant responses. The single-modality CSV agent showed stronger performance in basic and qualitative analysis, demonstrating the ability to link multiple CSV files and execute moderately complex processing steps. However, its inability to integrate non-CSV data modalities, such as LiDAR, prevented it from addressing quantitative analysis questions requiring multi-modal correlations.

In contrast, our proposed agent excelled across all levels of analysis. It effectively aligned and integrated data from diverse modalities, utilizing alignment mechanisms such as geo-information alignment for CSV and LiDAR integration, and unique identifiers for linking LiDAR and image datasets. This capability enabled it to answer complex questions, such as assessing land use changes in specific parks over time, which other agents were unable to accomplish.

In summary, the results demonstrate the significant limitations of existing models and agents in handling complex, multi-modal urban park development monitoring tasks. The data alignment mechanism in our framework and processing capabilities of our agent make it a robust and flexible solution, meeting the diverse analytical requirements of intelligent urban park development monitoring.

V. CONCLUSION

This study presents a multi-modal LLM agent framework tailored for urban park development monitoring, addressing the limitations of traditional change detection methods in high-level and intelligent analysis by leveraging the semantic understanding and reasoning capabilities of LLMs. Moreover, the proposed framework integrates a horizontal and vertical data alignment mechanism to ensure consistency and effective multi-modal data tracking. Additionally, a domain-specific toolkit mitigates hallucination issues, enhancing reliability. Experimental results demonstrate that our approach outperforms existing models, enabling robust and scalable multi-modal information fusion and analysis. This work provides a significant step forward in meeting the evolving demands of urban park development monitoring and urban planning.

REFERENCES

[1] Loughran, Kevin. "Urban parks and urban problems: An historical perspective on green space development as a cultural fix." Urban Studies 57.11 (2020): 2321-2338.

[2] Asokan, Anju, and J. J. E. S. I. Anitha. "Change detection techniques for remote sensing applications: A survey." Earth Science Informatics 12 (2019): 143-160.

[3] Hurst, Aaron, et al. "Gpt-4o system card." arXiv preprint arXiv:2410.21276 (2024).

[4] Lu, Dengsheng, et al. "Change detection techniques." International journal of remote sensing 25.12 (2004): 2365-2401.

[5] Parelius, Eleonora Jonasova. "A review of deep-learning methods for change detection in multispectral remote sensing images." Remote Sensing 15.8 (2023): 2092.

[6] Xiao, Zixuan, et al. "Few-shot object detection with self-adaptive attention network for remote sensing images." IEEE Journal of Selected Topics in Applied Earth Observations and Remote Sensing 14 (2021): 4854-4865.

[7] Xiao, Zixuan, et al. "Few-shot object detection with feature attention highlight module in remote sensing images." 2020 International Conference on Image, Video Processing and Artificial Intelligence. Vol. 11584. SPIE, 2020.

[8] Wei, Jason, et al. "Chain-of-thought prompting elicits reasoning in large language models." Advances in neural information processing systems 35 (2022): 24824-24837.

[9] Brown, Tom, et al. "Language models are few-shot learners." Advances in neural information processing systems 33 (2020): 1877-1901.

[10] Ge, Yuying, et al. "Making llama see and draw with seed tokenizer." arXiv preprint arXiv:2310.01218 (2023).

[11] Zhang, Dong, et al. "Speechgpt: Empowering large language models with intrinsic cross-modal conversational abilities." arXiv preprint arXiv:2305.11000 (2023).

[12] Wu, Chenfei, et al. "Visual chatgpt: Talking, drawing and editing with visual foundation models." arXiv preprint arXiv:2303.04671 (2023).